\documentclass[fleqn,10pt]{wlscirep}
\usepackage[utf8]{inputenc}
\usepackage[T1]{fontenc}
\usepackage{textcomp}

\usepackage{amsfonts}
\usepackage{amsmath}
\usepackage{amssymb}
\usepackage{amsthm}
\usepackage{multicol}
\usepackage{multirow}
\usepackage{subfig}
\usepackage[labelfont=bf,textfont=bf]{caption}
\usepackage{url}
\usepackage{xcolor}
\usepackage{hyperref}

\usepackage{array}
\usepackage{makecell}
\usepackage{amsmath,amssymb,amsfonts}
\usepackage{tabularx}
\usepackage[export]{adjustbox}
\usepackage[algo2e]{mathtools,algorithm2e}

\usepackage{mathtools}
\usepackage{siunitx}
\usepackage[capitalise,nameinlink,noabbrev]{cleveref}
\usepackage{booktabs}
\usepackage{bbm}
\hypersetup{
 colorlinks=true,
 linkcolor=blue,
 urlcolor=cyan,
 citecolor=black
}

\newcommand{\etal}{~\textit{et al.}}

\newcommand{\croppdf}[1]{\IfFileExists{#1-crop.pdf}{}{\immediate\write18{pdfcrop #1.pdf}}}
\DeclareMathOperator*{\argmax}{arg\,max}

\newcommand{\supplementarysection}{%
\newpage
  \setcounter{figure}{0}
  \let\oldthefigure\thefigure
  \renewcommand{\thefigure}{S\oldthefigure}

  \setcounter{table}{0}
  \let\oldthetable\thetable
  \renewcommand{\thetable}{S\oldthetable}

  \setcounter{section}{1}
   \let\oldthesubsection\thesubsection
   \renewcommand{\thesubsection}{S\oldthesubsection}
  \section*{Supplementary material}
}

\title{Toward Cost-efficient Adaptive Clinical Trials in Knee Osteoarthritis with Reinforcement Learning}

\author[1]{Khanh Nguyen}
\author[1]{Huy Hoang Nguyen}
\author[1]{Egor Panfilov}
\author[1, *]{Aleksei Tiulpin}
\affil[1]{Research Unit of Health Sciences and Technology, University of Oulu, Oulu, Finland}

\affil[*]{Corresponding author: aleksei.tiulpin@oulu.fi}


\begin{abstract}
Osteoarthritis (OA) is the most common musculoskeletal disease, with knee OA (KOA) being one of the leading causes of disability and a significant economic burden. Predicting KOA progression is crucial for improving patient outcomes, optimizing healthcare resources, studying the disease, and developing new treatments. The latter application particularly requires one to understand the disease progression in order to collect the most informative data at the right time. Existing methods, however, are limited by their static nature and their focus on individual joints, leading to suboptimal predictive performance and downstream utility. Our study proposes a new method that allows to dynamically monitor patients rather than individual joints with KOA using a novel Active Sensing (AS) approach powered by Reinforcement Learning (RL). Our key idea is to directly optimize for the downstream task by training an agent that maximizes informative data collection while minimizing overall costs. Our RL-based method leverages a specially designed reward function to monitor disease progression across multiple body parts, employs multimodal deep learning, and requires no human input during testing. Extensive numerical experiments demonstrate that our approach outperforms current state-of-the-art models, paving the way for the next generation of KOA trials. The codes behind our work are publicly available at \url{https://github.com/imedslab/OACostSensitivityRL}.
\end{abstract}

\begin{document}

\flushbottom
\maketitle
\thispagestyle{empty}

\section*{Introduction}

Osteoarthritis (OA) is a degenerative joint disorder that primarily affects cartilage and results in its gradual loss and subsequent damage to the joint~\cite{primorac2020knee}. Knee OA (KOA) is the most frequent type of OA and it is the focus of this work. The most common symptoms of KOA include pain, stiffness, and reduced joint flexibility. All of these significantly impact an individual's quality of life, and as the condition progresses, symptoms often worsen, leading to disability~\cite{mcalindon1993determinants}. 

To date, there are no effective treatments that can prevent permanent joint damage and disability due to KOA~\cite{schafer2022targeted}. In the final stages of KOA, a costly and highly invasive intervention -- a total knee replacement (TKR) -- is performed to improve the patient's health and well-being. TKR surgery can cost up to \$50,000 per patient~\cite{price2018knee,evans2018does,phillips2019much}, and is reported to produce unsatisfactory results in 15-20\% of cases~\cite{marsh2022health, bourne2010patient,price2018knee}. Due to the high prevalence of KOA and rapid population aging, TKR gradually becomes a financial burden not only for individuals~\cite{kjellberg2016nationwide} but also for the healthcare system~\cite{gandhi2023costs}. Therefore, KOA requires next-generation disease-modifying drugs (DMOADs)~\cite{oo2018disease,rodriguez2023current}, which can slow down the progression of the disease, thereby delaying the need for TKR~\cite{latourte2020emerging,cho2021disease}. 

Although the research community has been working on DMOADs in KOA for years~\cite{ITIC,hunter2011pharmacologic,schafer2022targeted}, no such treatments have yet been found to be effective~\cite{rodriguez2023current}. One of the challenges in clinical trials developing DMOADs for KOA is the long patient follow-up time~\cite{latourte2020emerging}. The etiology of KOA is poorly understood and it is often a slowly progressing disease~\cite{driban2020risk}. Thus, many subjects recruited into KOA trials and cohorts do not develop the disease at all, and some develop only early signs of the disease at the end of a study. Outside of KOA, adaptive methods for data collection in modern clinical trials~\cite{spreafico2021future,cuzick2023importance} are becoming instrumental in providing better, more efficient, and effective patient monitoring compared to routine scheduling and randomized participant selection~\cite{wu2022optimizing,yala2022optimizing}. Personalized disease modeling is essential to enable such trials and new treatments, as a ``one-size-fits-all'' approach has low chances of success in OA~\cite{schafer2022targeted}. 

Despite recent advances in KOA progression modeling~\cite{halilaj2018modeling,mccabe2022externally,tiulpin2019multimodal,nguyen2022climat,nguyen2023clinically}, translating existing models into real-world applications is challenging for several reasons. First, the financial downstream impact of the disease progression models is difficult to assess during the model development phase. As such, current models do not account for the \textit{follow-up costs} and possible \textit{future expenses} associated with the progression of structural KOA when estimating whether a person will develop KOA in the future. Secondly, existing models~\cite{tiulpin2019multimodal,tiulpin2022predicting,nguyen2023clinically,hirvasniemi2023knee,hu2022adversarial,panfilov2022predicting,panfilov2023end} focus primarily on knee-level progression events and overlook the broader patient-level context of KOA, where progression may occur concurrently in both knees. Third, the severity of KOA can increase multiple times during a study or trial, and there are uncertain factors that can alter the course of the disease (for example, future injuries). Therefore, long-term prediction is challenging and a different approach is needed to predict the progression of OA.

Active Sensing (AS)~\cite{yu2009active} addresses the question of \textit{when} to follow up the patient and with \textit{what tools and modalities}. In the case of KOA, one would want an AS policy to minimize follow-up costs while maximizing the chances of capturing patient-level progression. However, solving such a problem is challenging for a na\"ive supervised learning (SL) approach used to train disease progression models~\cite{nguyen2023clinically,panfilov2023end,tiulpin2019multimodal}. This is due to AS making predictions to acquire new data that later inform subsequent predictions, leading to a continuous cycle of adaptive data collection and decision-making under uncertainty. This task, however, can be solved with Reinforcement Learning (RL)~\cite{sutton2018reinforcement}.  

In this work, we porpose an AS methodology for KOA using RL, unlocking next-generation clinical trials and study cohorts to understand the progression of KOA and develop new treatments. We use RL to make sequential decisions based on the interactions between a decision-making agent and an environment. Here, the environment simulates a clinical trial or another setting in which patients need to be observed at intervals for a prolonged period of time. The specific contributions of our work are: 
\begin{itemize}
\itemsep0em 
    \item We introduce a formal framework for personalized AS into the KOA domain to improve the efficiency of clinical trials.
    \item We employ an RL-based model and develop a novel reward function to derive a personalized patient follow-up schedule that is being dynamically refined. This reward function unifies the costs and utility of data collection, allowing financial planning of real-world experiments.
    \item We perform an extensive experimental analysis and show how the developed method behaves in different settings.
\end{itemize}

To foster real-world impact, we openly release the code and a newly developed RL environment, paving the way for the next generation of clinical trials and ways to collect more informative datasets in KOA and beyond.

\section*{Results}
\begin{figure*}[!ht]
    \centering
    \croppdf{figures/PLASO_RL}
    \includegraphics[width=\textwidth]{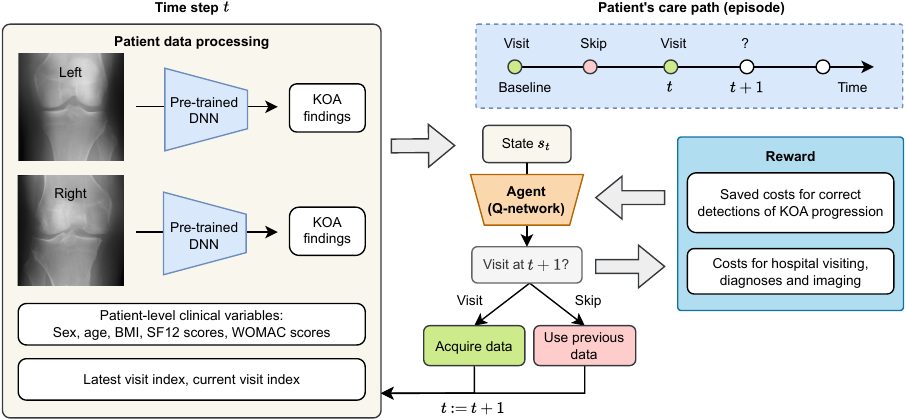}
    \caption{The workflow of our active sensing method, which performs decision-making under uncertainty. The state at each time point is associated with the data acquired at the latest patient visit. The reward function is designed to maximize the efficiency of hospital visits by taking into account radiographic changes and hospital visit costs. The set of actions comprises two elements -- follow-up at time $t$ or skip. DNN = deep neural network; KOA = knee osteoarthritis; BMI = body mass index; SF12 = 12-item Short Form Survey; WOMAC = the Western Ontario and McMaster Universities Osteoarthritis Index.}
    \label{fig:workflow}
\end{figure*}

\paragraph{Our proposed Active Sensing Method}
The overview of our method is shown in~\Cref{fig:workflow}. We designed our RL agent to autonomously follow patients to maximize the number of progression detections while minimizing overall costs. The RL agent is a Q-Network~\cite{sutton2018reinforcement} that processes the state $\mathbf{s}_t$ revealed by the environment (that is, the status of the KOA) and predicts the value of a follow-up or dismiss action.

The RL agent was trained in an episodic setting, where each episode represented the subject's participation in an observational study. In this work, we used the openly available Osteoarthritis Initiative (OAI; \url{https://nda.nih.gov/oai/}) data set to simulate the real-world setting. 

At each step, the agent (AS policy) was optimized by the reward function, which was specifically designed for the setting of AS, and aimed to minimize overall costs (measured in dollars) while maximizing the number of follow-up examinations where the disease has progressed. We utilized the fixed Joint Space Width @ 250 (fJSW) imaging biomarker, which is measured from X-ray images~\cite{duryea2010comparison}. The reward function is one of the main results of the present work, which we explain below and demonstrate its validity through numerical experiments. 

\paragraph{Reward function}
Assume we are provided with a set of progression events $\mathbf{M}_p=\{t_j\}_{j=1}^{N_p}$, where $N_p$ is the total number of progression events. The details on the exact definitions of this set are shown in Methods. For each subject, at time $t \geq 0$, an agent takes one of two actions: follow-up or dismiss (that is, skip examination) at the next time point $t+1$. The rewards for these actions are denoted as $r_f$ and $r_d$, respectively. The reference point $t_r$ is updated with the latest visit, ensuring that the reward calculations are based on the most recent state.

By comparing $t+1$ and $t_p$ (the closest progression time point extracted from the patient-level progression set $\mathbf{M}_p$), we derive five possible scenarios. Specifically, there are three cases for the correctness of the follow-up action ($a=1$): \emph{early visit}, \emph{timely visit}, and \emph{late visit}, and two cases for the dismissal action ($a=0$): \emph{true dismissal} and \emph{false dismissal}. The reward is expressed as $r_t(\mathbf{s}_t, a) = [a = 1]r_f + [a = 0]r_d$, having 
\begin{align}
    r_{f} &= \left\{ 
    \begin{matrix*}[l]
    \phantom{-}e^{- \tau(t)} \Delta(t_r, t+1) - \lambda & \,\, \mathrm{if} \,\,\, t+1 < {t_{p}} \ \textit{(early visit)} \\  \\ 
    \phantom{-}\Delta(t_r, t + 1) - \lambda & \,\, \mathrm{if} \,\,\, t+1 = {t_{p}} \ \textit{(timely visit)} \\ \\
    \ -\alpha \  e^{\tau(t)} \Delta({t_r, t_p}) - \lambda & \,\, \mathrm{if} \,\,\, t+1 > {t_{p}} \ \textit{(late visit)}
    \end{matrix*}\right. \label{eq:reward_fl}
\end{align}
\begin{align}
    r_{d} &= \left\{ 
    \begin{matrix*}[l]
    \phantom{-}\ \beta  & \,\, \mathrm{if} \ t+1 < {t_{p}} \ \textit{(true dismissal)} \\ \\
     - e^{\tau(t)} \Delta({t_r,t_p}) & \,\, \mathrm{if} \ t+1 \geq {t_{p}} \ \textit{(false dismissal)},
    \end{matrix*}\right. \label{eq:reward_dismiss}
\end{align}
where $\lambda \in \mathbb{R}^+$ is the fixed cost of data acquisition, and $\Delta(t_r, \cdot)$ is the data acquisition utility function, representing the cost of significant changes relative to $t_r$:
\begin{align}
    \Delta(t_r, \cdot) = \delta(t_r, \cdot)\left[\delta(t_r, \cdot) \geq c\kappa\right].
\label{eq:Delta-func}
\end{align}

The reward for a late visit is equivalent to that for a false dismissal, but is discounted by $\alpha \in [0,1)$ and decreased by $\lambda$. In addition, $\beta < c\kappa$ is the positive reward for the correct dismissal action. In our notation, $\kappa$ represents the minimum detectable significant change in fJSW, and $c$ represents the monetary cost of a 1mm loss in articular cartilage (MLAC). Having $\kappa$ is important since fJSW has limited repeatability and is affected by measurement noise, which can arise from factors such as X-ray positioning. 

To incorporate the difference between the time of making decision $t$ and the actual KOA progression timepoint $t_p$ into the reward function, we introduce the modulating function $\tau: [0, T] \rightarrow [0, 1]$, defined as:
\begin{align}
    \tau(t) = \frac{|t+1 - t_p|}{T}.
\end{align}
We consider two specific cases that require such a modulating function: early follow-up and late follow-up/false dismissal. For the first case, we use an exponential decay function $e^{-\tau(t)}$. Therefore, increasing $\tau(t)$ reduces the reward that an agent can obtain. For the second case, we penalize late or false dismissal actions and use the term $-e^{\tau(t)}$. As $\tau(t)$ increases, the negative reward increases, preventing the agent from delaying necessary actions or incorrectly dismissing the patient.

\subsection*{Reward function ablation}
We performed an ablation study on the parameters $\alpha$ and $\beta$ in our reward function. These parameters correspond to the \textit{late visit} and \textit{true dismissal} actions, respectively. Our experiment involved an evaluation of the results on a grid of hyperparameters of the reward, where $\alpha \in \{0.25,\ 0.50,\ 0.75,\ 1.00\}$ and $\beta \in \{0.1c,\  0.3c,\ 0.5c,\ 0.7c\}$. 

\begin{figure}[!ht]
\centering

\IfFileExists{figures/results/Alpha_beta_BA-crop.pdf}{}{\immediate\write18{pdfcrop
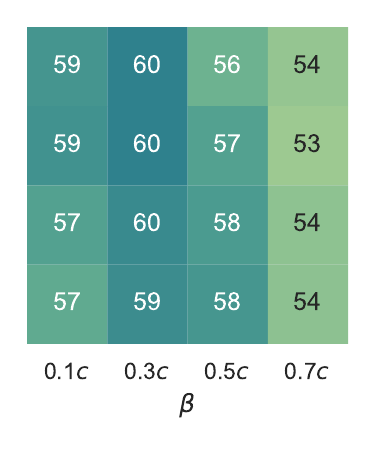}}
\IfFileExists{figures/results/Alpha_beta_recall_over_cost-crop.pdf}{}{\immediate\write18{pdfcrop
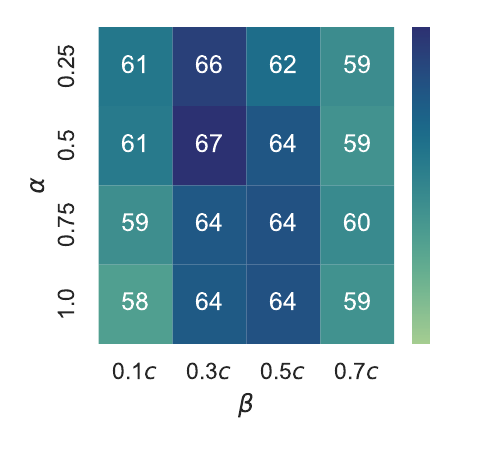}}
\subfloat[\centering Recall-over-cost ratio \label{fig:recall_over_cost_alpha_beta}]{
\includegraphics[height=0.3\linewidth]{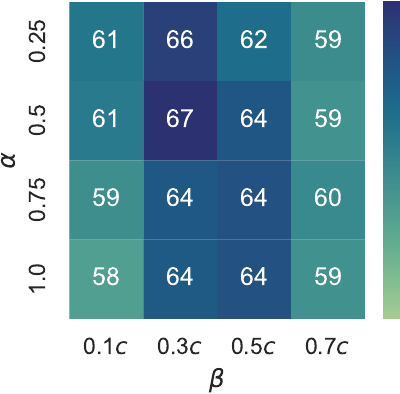}}
\hspace{5em}
\subfloat[\centering Balanced accuracy \label{fig:BA_alpha_beta}]{
\includegraphics[height=0.3\linewidth]{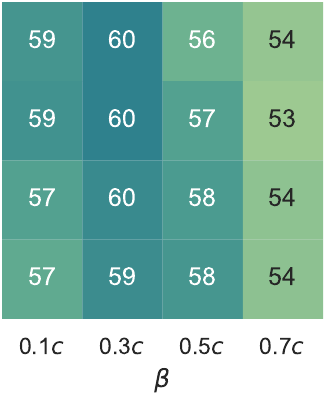}}
\caption{Ablation study of parameters $\alpha$ and $\beta$, evaluated using ratio of recall over acquisition cost and balanced accuracy. The optimal values of $\alpha$ and $\beta$ are chosen based on these metrics.}
\label{fig:Scanning_alpha_beta}
\end{figure}

Given that the reward scale fluctuates with adjustments to these parameters, as indicated in~\Cref{eq:reward_dismiss,eq:reward_fl}, we introduced the normalized score: recall-over-cost ratio as a metric to evaluate the results. In addition, we used the balanced accuracy (BA) score. The results are illustrated in~\Cref{fig:Scanning_alpha_beta}. Our findings revealed that the optimal parameter combination was $\alpha=0.5$ and $\beta=0.3c$. This combination yielded a BA of $60\%$ and a recall-over-cost ratio of $67\%$, the highest scores among other combinations. 

\subsection*{Training process convergence in an episodic setup}
\begin{figure}[ht!]
\IfFileExists{figures/results/Train_curve_varied_eps_gamma_1pic-crop.pdf}{}{\immediate\write18{pdfcrop
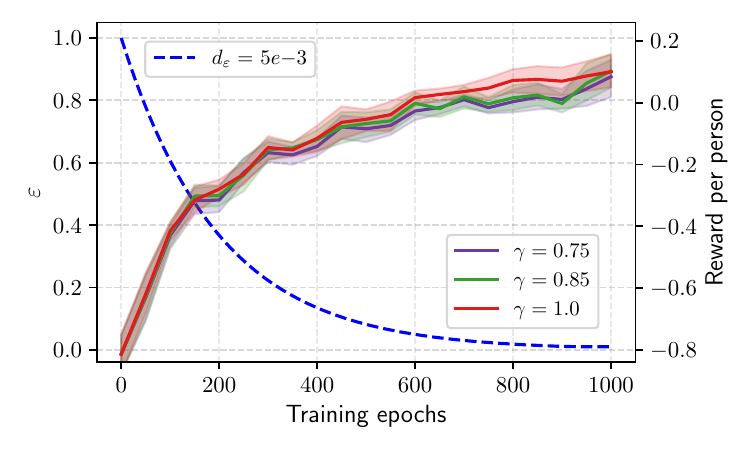}}
\centering
\includegraphics[width=0.6\linewidth]{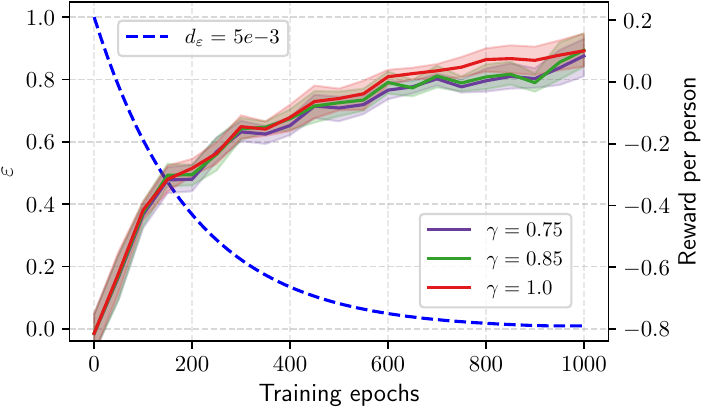}
\caption{Convergence of model training. We adjusted the balance between exploration and exploitation by decaying $\varepsilon$ over epochs with different discount factors $\gamma$. The dashed line indicates the decay of $\varepsilon$ (left y-axis). The color lines indicate the improvement of RPP over 4 years under RL-based policy (right y-axis). Colors correspond to different discount factors $\gamma \in \{0.75,0.85,1.0\}$. The policy trained with $\gamma=1.0$ was the most stable and achieved the highest reward over training.}
\label{fig:converg_decays_1pic}
\end{figure}

For the $\varepsilon$-greedy policy, we conducted the experiment where $d_\varepsilon$, represented the decay of exploration, and was selected from the set $\{1e-2$, $5e-3$, $1e-3\}$. Each setting was repeated with $10$ random seeds. With $3$ different $\varepsilon$-decayed curves, the training policies converged, yet at different training times. The convergence of training in any exploration-exploitation strategy serves as preliminary evidence of the effectiveness of our proposed reward function in identifying an optimal policy. By comparing the learning curves of three $\varepsilon$-greedy policies over $1000$ epochs, we concluded that $d_\varepsilon=5e-3$ is the most appropriate value to define the exploration-exploitation process within an acceptable training time (see~\Cref{fig:converg_decays_1pic}).

In addition to the above, an ablation study was performed to investigate the impact of discount factor $\gamma$ on learning progress (\Cref{fig:converg_decays_1pic}). The results showed that the values of $\gamma < 1.0$ to discount future rewards yielded lower performance. Thus, $\gamma = 1$ was chosen for all subsequent experiments.

\subsection*{Numerical comparisons to existing methods}
We compared our method to a wide range of reference methods belonging to three categories of sensing: random, routine, and active sensing. The quantitative results are presented in~\Cref{tbl:baselines_comparision}. The random policy was parameterized by the probability of taking follow-up action. Here, we evaluated four settings: $0.3$, $0.5$, $0.7$, and $0.9$.
The set of routine policies comprised ``no sensing'' (NOS), annual sensing (ANS), and biannual sensing (BANS). Dynamic policies included logistic regression (LR), Cox regression (Cox), recurrent neural network (RNN)-based methods, and transformer-based CLIMATv2~\cite{nguyen2022climat,nguyen2023clinically}. More details on these baselines are provided in \Cref{supp:baselines}. All experiments were run 10 times with different random seeds, and we present the average scores over these runs. Standard errors are omitted in cases where they were less than $0.01$.

\begin{table}[ht!]
\centering
\caption{Comparison between our Reinforcement Learning-based policy and the baseline approaches in sensing KOA progression. The methods are compared by individual gain (RPP; reward per person), acquisition costs, average balanced accuracy (BA), and average recall. We categorized all policies under $3$ types: random, routine, and active.
($\star$) indicates a binomial distribution with $n=4$. The highest overall score for each metric is underlined. The top scores among active sensing policies are highlighted in bold.
}\label{tbl:baselines_comparision}

\begin{tabular}{llcccc}
\toprule
\textbf{Setting} & \textbf{Method}  & \textbf{RPP $\uparrow$} & \textbf{Acquisition cost $\downarrow$} & \textbf{Average BA $\uparrow$} & \textbf{Average recall $\uparrow$} \\
\midrule 
\multirow{4}{*}{Random$^{\star}$} & $p_{(a=1)}=0.3$  & -0.78 $_{\pm 0.06}$ & 0.59 $_{\pm 0.01}$ & 50.42 $_{\pm 0.49}$ & 30.15 $_{\pm 0.87}$  \\
& $p_{(a=1)}=0.5$  & -0.65 $_{\pm 0.05}$ & 1.00 $_{\pm 0.01}$ & 49.69 $_{\pm 0.46}$ & 49.48 $_{\pm 0.88}$  \\
 & $p_{(a=1)}=0.7$   & -0.56 $_{\pm 0.04}$ & 1.40 $_{\pm 0.01}$ & 50.73 $_{\pm 0.51}$ & 71.15 $_{\pm 0.96}$ \\
 & $p_{(a=1)}=0.9$   & -0.72 $_{\pm 0.01}$ & 1.79 $_{\pm 0.01}$ & 50.28 $_{\pm 0.34}$ & 90.09 $_{\pm 0.60}$  \\
\midrule
\multirow{3}{*}{Routine} & NOS  &  -1.64   &  \underline{0}         & 50          &   0        \\
& BANS   &  -0.10   & 1           & 50          & 50       \\
& ANS   & -0.86          & 2            & 50         & \underline{100}         \\
\midrule
\multirow{9}{*}{Active}& LR  &-0.67 $_{\pm 0.03}$ & 0.73 $_{\pm 0.01}$ & 50.89 $_{\pm 0.57}$ & 38.24 $_{\pm 0.94}$ \\
&Cox  &-0.86 $_{\pm 0.01}$ & 0.40 $_{\pm 0.01}$ & 53.80 $_{\pm 0.01}$ & 19.27 $_{\pm 0.01}$ \\
&DeepHit  &-0.30 $_{\pm 0.01}$ & 1.18 $_{\pm 0.01}$ & 52.39 $_{\pm 0.01}$ & 62.98 $_{\pm 0.01}$\\
&GRU  &-1.23 $_{\pm 0.10}$ & 0.24 $_{\pm 0.07}$ & 52.79 $_{\pm 0.45}$ & 17.00 $_{\pm 4.22}$ \\
&LSTM  & -1.34 $_{\pm 0.08}$  & \textbf{0.12  $_{\pm 0.02}$} & 52.30  $_{\pm 0.74}$ & 9.74 $_{\pm 2.43}$ \\
&CLIMATv2  & -0.54 $_{\pm 0.04}$ & 0.53 $_{\pm 0.04}$ & 56.41 $_{\pm 0.41}$ & 37.78 $_{\pm 2.27}$ \\
\cmidrule{2-6}
& RL (Ours)    & \textbf{\underline{0.20 $_{\pm 0.03}$}}          &  1.11 $_{\pm 0.02}$           & \textbf{\underline{60.33 $_{\pm 0.32}$}}       & \textbf{73.67 $_{\pm 1.00}$}             \\
\bottomrule
\end{tabular}
\end{table}

In general, random policy could not exceed BA of $51\%$, regardless of the probability of selecting the follow-up action. As the probability of taking an action increased from $30\%$ to $90\%$, the trend of the average rewards per person (RPP; see definition in Methods) exhibited a parabolic shape, reaching its peak near $70\%$. The random policy settings of $30\%$ and $90\%$ were approximately equal to each other in RPP, which were the lowest among the random policies evaluated.

Among the three routine sensing approaches, the BANS policy yielded the highest reward of $-0.1$, which was still negative. The NOS policy resulted in the lowest reward not only among the three, but also across \textit{all} methods. 
The ANS policy, a standard method of data collection that ensures that there are no missed follow-up visits, yielded the highest acquisition cost in comparison to other methods. Compared to this policy, our method was able to identify $73.67\%$ of the progression events while reducing the acquisition cost by up to $45.5\%$. 

In general, AS baselines tended to improve BA at the expense of lower recall when compared to the BANS policy. Specifically, the improvement in BA ranged from $0.02\%$ to $6.41\%$, with the most significant increase observed with CLIMATv2. However, DeepHit was the only AS reference that improved recall compared to the BANS policy, demonstrating an increase of $12.98\%$. As a result, DeepHit achieved the highest reward among all baselines, although its total reward was still negative.

Unlike AS baselines, our RL-based policy led to improvements in BA, recall, and RPP. Specifically, our method significantly outperformed CLIMATv2 with increases of $3.9\%$ in BA and $35.9\%$ in recall score. Compared to DeepHit, the RL-based method demonstrated substantial improvements of $7.9\%$ in BA and $10.7\%$ in recall score. 

\subsection*{Impact of financial factors on decision-making}
As the cost-related parameters $\lambda$ and $c$ are key factors in our reward function, our aim was to quantitatively assess their impacts on the method performance. We conducted two experiments that involved our proposed method, CLIMATv2, ANS, and NOS policies. In the first experiment, we varied the hospital visit cost $\lambda$, while keeping the MLAC $c$ unchanged. In the second experiment, we inverted the setting.

\begin{figure*}[!ht]%
\centering
\IfFileExists{figures/results/adjust_lambda-crop.pdf}{}{\immediate\write18{pdfcrop
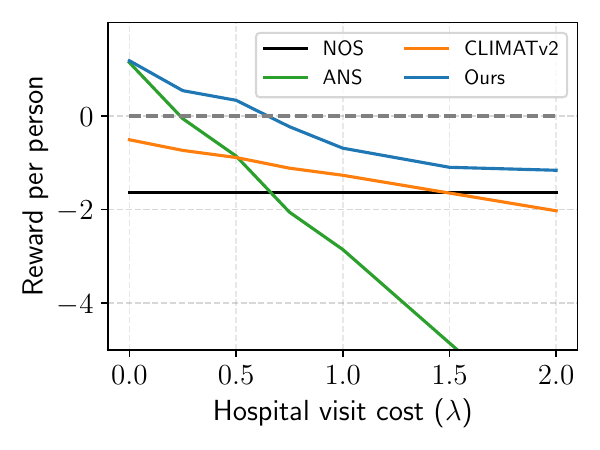}}
\IfFileExists{figures/results/adjust_c-crop.pdf}{}{\immediate\write18{pdfcrop
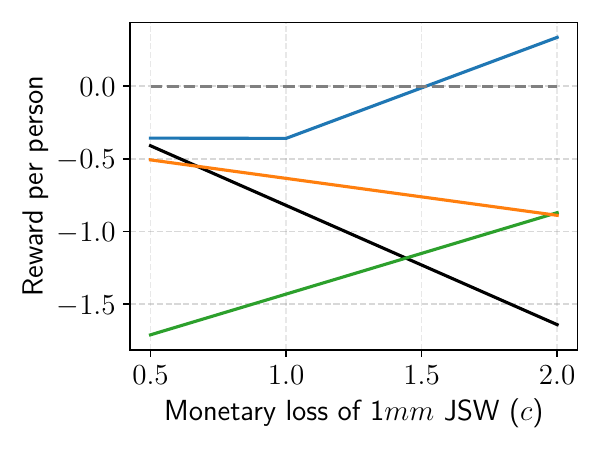}}
\IfFileExists{figures/results/reward_vs_cost_ratio-crop.pdf}{}{\immediate\write18{pdfcrop
figures/results/reward_vs_cost_ratio.pdf}}
\IfFileExists{figures/results/reward_vs_cost_ratio2-crop.pdf}{}{\immediate\write18{pdfcrop
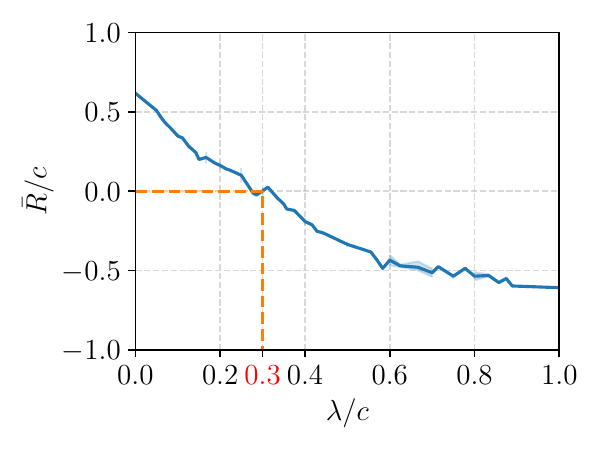}}
\subfloat[\centering With $c=2$, varying $\lambda$\label{fig:ajust_lambda}]{\includegraphics[height=0.23\linewidth, valign=t]{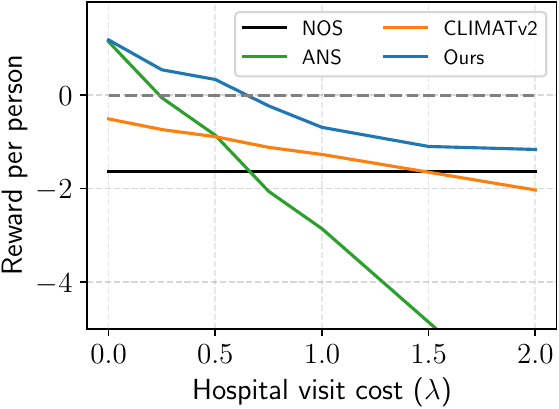} }%
\hfill
\subfloat[\centering With $\lambda=0.5$, varying $c$\label{fig:adjust_c}]{\includegraphics[height=0.23\linewidth, valign=t]{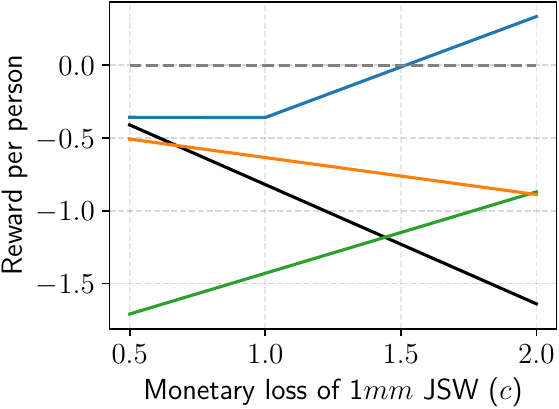}}%
\hfill
\subfloat[\centering Cost ratio and normalized reward \label{fig:relation_cost_ratio}]{\includegraphics[height=0.23\linewidth, valign=t]{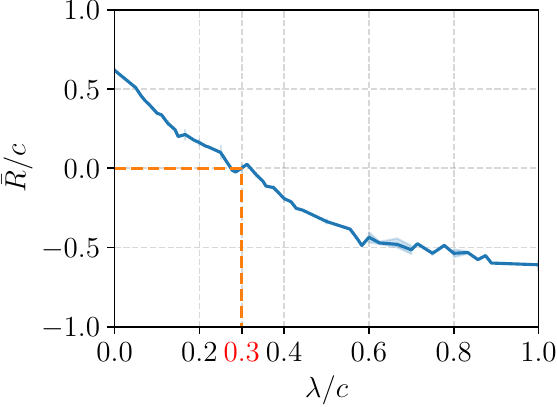}}
\caption{The reward-per-person (RPP) of ``no sensing'' (NOS), annual sensing (ANS), CLIMATv2, and our policy (``Ours''). Subgifure (a) shows how RPPs change with the hospital visit cost varying from \$0 to \$2,000. Subfigure (b) shows RPP changes with the varying cost of 1\textit{mm} fJSW degeneration from \$500 to \$2,000, which also corresponds to the increase of TKR cost. Subfigure (c) shows the relationship between the cost ratio ($\lambda/c$) and fJSW reward ($r/c$) for our RL-based method. The parameters were adjusted in both training and testing (i.e., deployment) subsets. We found that the ratio $\lambda/c$ of $0.3$ is a threshold for a cost-efficiency policy (highlighted in red).
}%
\label{fig:adjust_cost}%
\end{figure*}

We present the results of the first experiment in~\Cref{fig:ajust_lambda} (see~\Cref{tab:fig4a} for the detailed results), demonstrating the association between $\lambda$ and the RPP over $4$ year, with a fixed $c$ of $2$. Generally, all methods exhibited a decreasing trend in the reward as the hospital visit cost increased. Compared to the baselines mentioned above, our RL-based method consistently outperformed the selected baselines in RPP across all $\lambda \in [0,2]$ (see~\Cref{fig:ajust_lambda}). The NOS policy resulted in a constant RPP of $-1.64$ since no follow-up data were acquired. In ANS policy, increasing $\lambda$ from $0$ to $1.5$ yielded a significant drop in RPP of approximately $6.0$. Our proposed method matched the performance of the ANS policy at $\lambda = 0$ and approached the NOS policy when $\lambda$ increased.

The results of the second experiment are demonstrated in~\Cref{fig:adjust_c} (see~\Cref{tab:fig4b} for the raw data). When the hospital visit cost was kept unchanged $\lambda = 0.5$, the performance of CLIMATv2 and the NOS policies showed a decreasing trend as the cost of the loss of fJSW $c$ increased. In contrast, we observed that the ANS policy increased generally in proportion to $c$. Our policy achieved a constant reward of $-0.25$ for $c\leq1.0$, and demonstrated a linear increase for $c>1.0$. In general, our policy yielded the highest reward among the methods at any value of $c \in [0.5, 2]$. For values of $c$ greater than $1.5$, our method stand out as the only approach that yields positive RPP.

To gain deeper insights into the trade-offs related to the cost parameters $\lambda$ and $c$ and their impact on the results, we investigated the relationship between the relative cost $\frac{\lambda}{c}$ and the resulting normalized RPP, $\frac{\bar{R}}{c}$. Specifically, we varied the hospital visit cost $\lambda$ across the range $\{0,0.25,0.5,0.75,1,1.5,2,2.5,3,3.5,4\}$, while adjusting the MLAC $c$ within the set $\{0.5,1,1.5,2,3,4\}$. The results are illustrated in~\Cref{fig:relation_cost_ratio}. We observed that $\frac{\bar{R}}{c}$ exhibited an inverse, non-linear relationship with the increase of $\frac{\lambda}{c}$. In other words, a decrease of $\frac{\lambda}{c}$ was beneficial for the resulting RPP. Specifically, when the ratio $\frac{\lambda}{c}$ declined to $0.3$, it became the critical trade-off point at which our policy began to yield positive RPPs. The raw experimental data behind~\Cref{fig:relation_cost_ratio} is shown in~\Cref{tab:fig4c}

\subsection*{Patient-level versus knee-level prediction}
The majority of prior studies on KOA progression prediction considered the development of the disease within a single knee. However, we argued that for improved health outcomes, it is essential to incorporate information from both knees into patient-level input. Therefore, we conducted an experimental comparison of two approaches -- knee-level versus patient-level -- using our proposed method and the state-of-the-art baseline CLIMATv2~\cite{nguyen2023clinically}.

As we aimed to perform the patient-level progression prediction to assist in a personalized follow-up plan, we still evaluated all policies at the patient level. In the knee-level approach, we implemented two individual agents to predict the degeneration of each knee side. The final decision in the evaluation step was made by combining the actions of two knee-side agents, following the rule that any progression in either knee triggers a progression event. 

\begin{table}[ht!]
\caption{Overall performance of our RL-based method and CLIMATv2 in knee-level and patient-level KOA progression approaches. The values correspond to mean metric scores over 10 runs.}
\centering
\scalebox{0.95}{
\begin{tabular}{llrcc}
\toprule
\textbf{Approach}  & \textbf{Method}        &       \textbf{Reward}       & \textbf{BA} & \textbf{Recall} \\
\midrule 
\multirow{2}{*}{Knee-level} &CLIMATv2 & -0.56 & 54.97 & 33.20 \\
& Ours & 0.18  & 56.60  & 53.32              \\
\midrule
\multirow{2}{*}{Patient-level} &  CLIMATv2 & -0.54 & 56.41 & 37.78   \\
& Ours & 0.20   & 60.33   & 73.67  \\
\bottomrule
\end{tabular}
}
\label{tbl:knee_vs_patient}
\end{table}

The overall performance of our proposed method and CLIMATv2 in knee-level and patient-level approaches is shown in \Cref{tbl:knee_vs_patient} (see~\Cref{tab:fig5} for the raw data). Notably, the patient-level approach outperformed the knee-level approach for both CLIMATv2 and the RL-based policy, with average improvements in BA of $1.44\%$ and $3.73\%$, respectively, and in recall of $4.58\%$ and $20.25\%$, respectively.
Both methods gained a $0.02$ increase in RPP when switching from a knee-level to a patient-level approach.

\begin{figure}[!ht]
\centering

\IfFileExists{figures/results/Patient_vs_knee_BA-crop.pdf}{}{\immediate\write18{pdfcrop 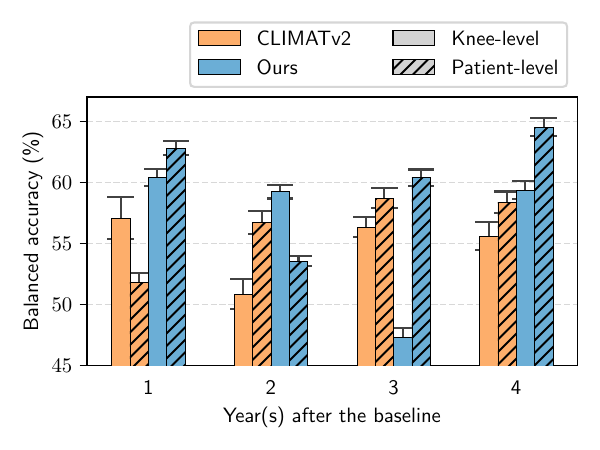}}
\IfFileExists{figures/results/Patient_vs_knee_Recall-crop.pdf}{}{\immediate\write18{pdfcrop 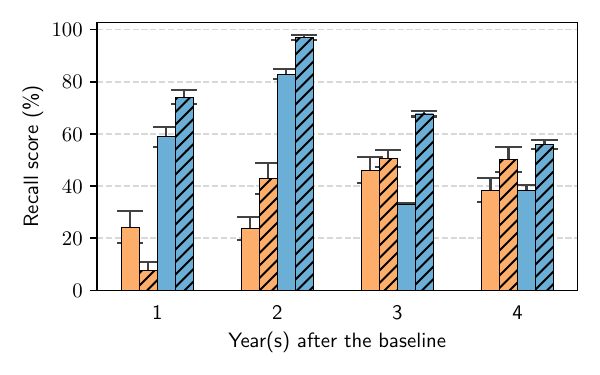}}

\subfloat[\centering Balanced accuracy]{
\includegraphics[width=0.45\linewidth]{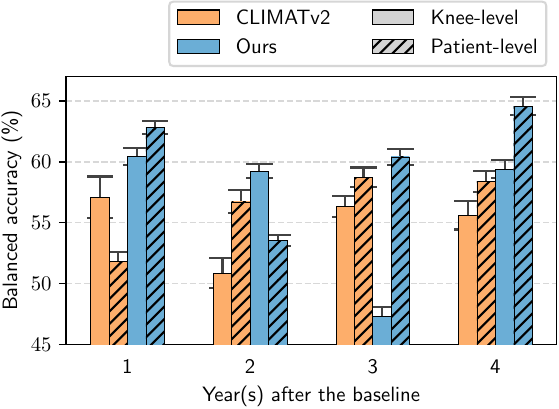}}~
\subfloat[\centering Recall]{\includegraphics[width=0.45\linewidth]{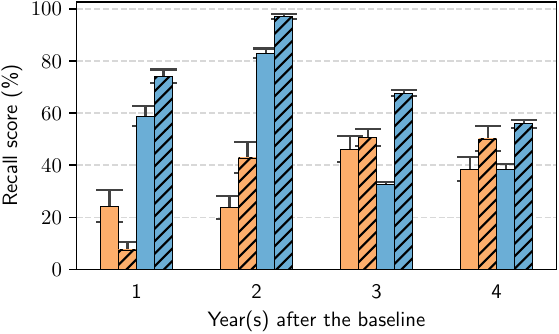}}

\caption{Performance metrics in knee-level and patient-level approaches at each follow-up year after the baseline visit. The error bars represent standard errors over 10 runs with random seeds.}
\label{fig:patient_vs_knee_metrics}
\end{figure}

To provide additional insights, we present the results for individual follow-ups in~\Cref{fig:patient_vs_knee_metrics}. Generally, patient-level models of both CLIMATv2 and our RL-based method showed slightly lower standard errors in BA and recall. At the one-year timepoint, our patient-level policy exceeded the patient-level CLIMATv2 with the highest difference of approximately 15\% in BA and 60\% in recall. Furthermore, the largest improvement from the knee-level to patient-level approach with our method occurred 3 years after the initial observation, which was approximately 20\% in BA and 40\% in recall. Overall, the patient-level approach with our method showed the highest BA and recall scores across the majority of the follow-ups.

\subsection*{The analysis of policy behavior}
Beyond the comparison with the baseline methods, we also investigated how the learned policy differs depending on the person's demographic info, symptomatic and radiographic status of KOA.

\begin{figure}[t]
\centering
\croppdf{figures/results/severity_vs_visits_lambda}
\includegraphics[width=0.6\linewidth]{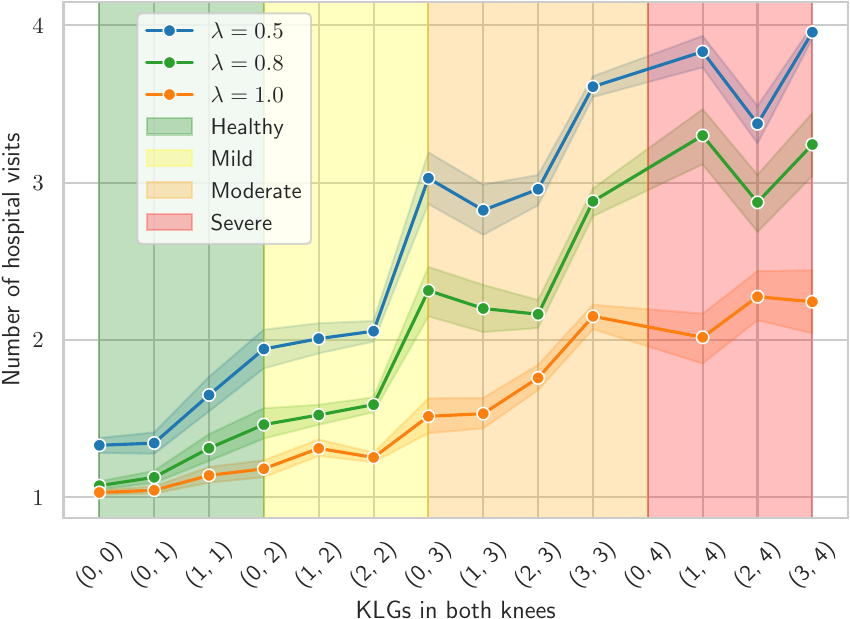}
\caption{Dependency between the baseline knee osteoarthritis (KOA) severity and the total number of follow-up decisions made with our method. Three RL-based policies obtained at different hospital visit costs $\lambda=\{0.5,0.8,1.0\}$ are shown. KOA severity in two knees is measured by Kellgren-Lawrence grades (KLGs) and arranged in an increasing manner along the x-axis. The numbers in parentheses indicate the KLGs without specifying the side. The color-coded regions denote patient-level KOA severity, where ``Healthy'' indicates neither knee has a KLG greater than 1, ``Mild'' -- greater than 2, ``Moderate'' -- greater than 3, and ``Severe'' indicates that at least one knee has KLG 4.
}
\label{fig:no_visits_severity}
\end{figure}

First, we evaluated the dependency between the number of follow-up decisions made by our method and the disease stage. The latter was defined by radiographic Kellgren-Lawrence grade (KLG) ranging from 0 to 4. \Cref{fig:no_visits_severity} visualizes the number of hospital visits per person according to the KLGs measured from the left and right knees at the initial visit (see~\Cref{tab:fig4} for the raw data). The KLG pairs were classified into $4$ subgroups: "Healthy" when neither knee had KLG greater than 1, "Mild" when neither knee had KLG greater than 2, "Moderate" when neither knee has KLG greater than 3, and "Severe" when at least one knee had KLG 4. We evaluated our method with $3$ different $\lambda$ -- $0.5$, $0.8$, and $1.0$. All of the resulting policies suggested at least one follow-up to all participants regardless of the costs. The results show that the method learned to be KOA severity-aware. Interestingly, the policies suggested more visits for patients with a more severe status of KOA, indicating that those patients are likely to progress faster. This observation has been consistent across all considered cost settings $\lambda$.

\begin{figure*}[t]

\IfFileExists{figures/results/Analyze_landscape_ver_bw2-crop.pdf}{}{\immediate\write18{pdfcrop
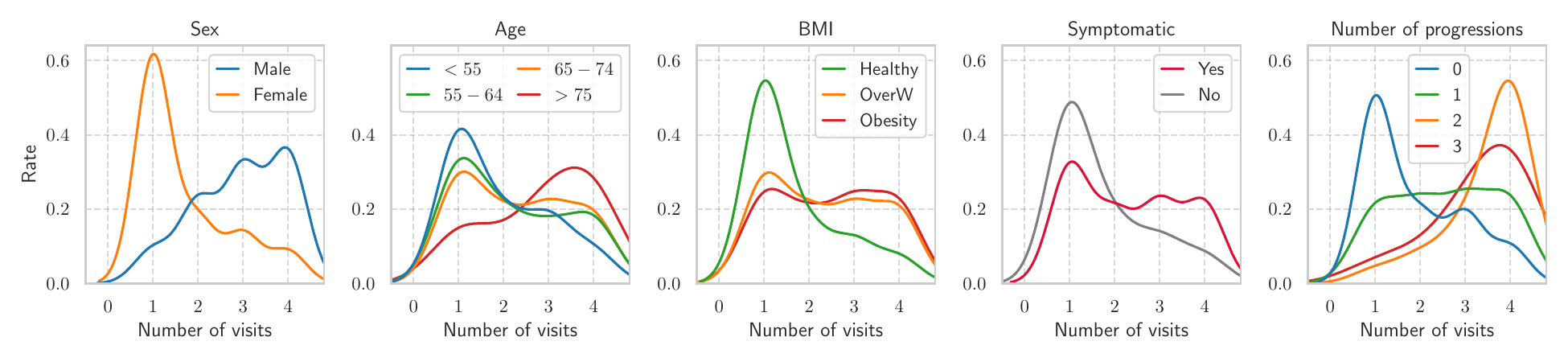}}
\includegraphics[width=1.00\linewidth]{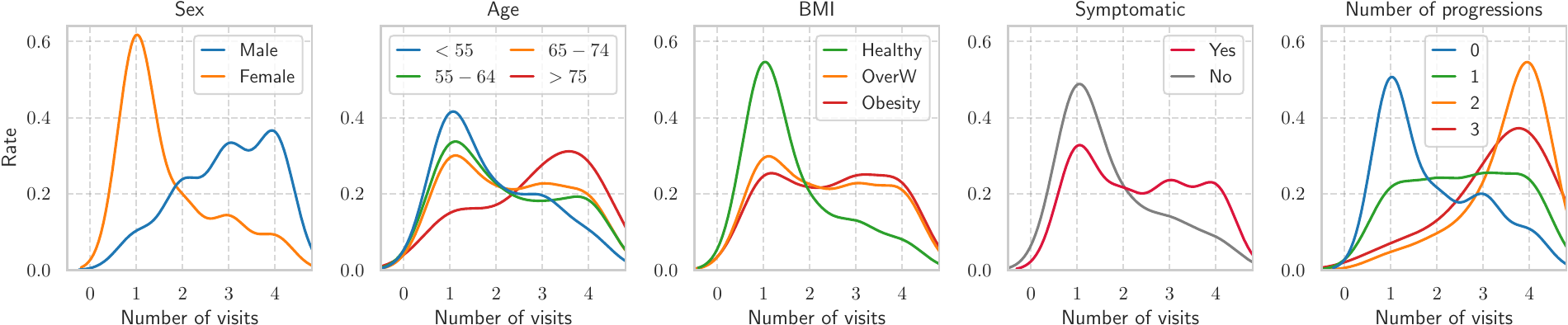}
\caption{The distribution of follow-up decision frequency (x-axis) in various patient groups with the visit cost of $\lambda=0.5$. Particularly, ``Age'', ``BMI'', and ``Symtomatic'' were classified at the baseline.}
\label{fig:frequency_category}
\end{figure*}

Finally, we also analyzed the frequency of follow-up decisions over 4 years in various patient groups defined based on sex, age, BMI, symptomatic KOA status, and the actual number of progression events (see ~\Cref{fig:frequency_category}). 

Our policy generally recommended at least one follow-up in the course of 4 years for all patients. A higher number of follow-up decisions were made for males compared to females. In terms of BMI, participants classified as overweight or obese obtained a higher number of follow-up decisions compared to those with normal BMI. Considering symptomatic status, our policy made more follow-up decisions for symptomatic patients than for asymptomatic ones, indicating its responsiveness to the presence of symptoms. In terms of patient age, the number of recommended follow-ups increased towards older participant groups, being the largest for those over $75$ years of age.
When we compared the frequency of suggested visits to the actual number of progression events, we found that participants with multiprogression were most likely to be sent for frequent visits. The method was most uncertain for patients with KLG 1.

\section*{Discussion}
In this study, we have developed an RL-based AS method that can be trained from cohort data. Our method allows to follow-up patients with KOA for radiographic changes with high sensitivity and cost-efficiency. 
The core of our methodology is an episodic Q-learning approach with a novel reward function. This function is specially designed to detect patient-level disease progression with recurrent events in an online AS setting.
Through extensive experiments, we have shown that our method outperforms various baselines in terms of balancing disease outcomes and costs, and it is the only method to achieve a positive RPP among those compared. To our knowledge, this work is the first to consider dynamic monitoring of the progression of KOA, and among the first to tackle the problem of optimal data collection in KOA.

Several radiographic imaging biomarkers, including the OARSI grade~\cite{altman2007atlas}, the KLG system~\cite{kellgren1957radiological}, and the JSW degeneration~\cite{duryea2010comparison}, are used to define the progression of KOA. These imaging biomarkers track the disease progression within a single knee joint. Our work argues that a policy for optimal data collection and disease tracking should consider the evolution of OA in both knee sides simultaneously. Although OA can develop in multiple joints, only a few studies so far focused on predicting progression at the patient level. Our study showed that an AS policy that tracks the progression of OA at the patient level substantially outperforms that at the knee level. To make RL work in the AS context, we have developed a novel reward function, which balances the agent's ability to detect the progression of KOA as well as the acquisition costs. The proposed reward incorporates follow-up and dismissal actions, divided into mutually exclusive scenarios. These naturally occur when making a decision to perform data acquisition. 

To address the challenge of assigning economical value to every action, we have introduced hyperparameters $\alpha$ and $\beta$ into our reward function. Parameter $\alpha$ mitigates the punishment for late follow-up on false dismissal (i.e., ``late better than none'') and $\beta$ represents the gain in a true dismissal reward. We propose that $\alpha$ and $\beta$ should be customized to better fit different economical settings and downstream applications, although investigating them is beyond the scope of this work. Our ablation experiments shown in \Cref{fig:Scanning_alpha_beta} provide insights into their selection procedure in the context of KOA.

Given the complexity of KOA progression, it is worth discussing various sensing policies that we implemented as baselines. Annual sensing, while thorough, is cost-ineffective as it incurs the highest acquisition costs. This is already known, as it served as the motivation for our study. The BANS policy seems to be an easy-to-implement and much better policy cost-wise, but it detects only $50\%$ of progressors. Implementing predictive models as a base for AS was expected to bring positive outcomes; however, the results did not yield a positive reward in any of the cases. In a general cost setting over $4$ years, to collect over $70\%$ of all progression events, the acquisition budget requires $\$1,400 - \$2,000$ for each participant, whereas our policy costs only around $\$1,110$, which translates into approximately $21\% - 45\%$ cost savings per patient in a trial or a cohort. 

Beyond the hyperparameters $\alpha$ and $\beta$, our policy's decision-making quality also depends on the ratio of acquisition cost to treatment cost ($\lambda / c$). We found that all settings where $\lambda / c\leq 0.3$ are the ones that give nonnegative rewards. 
This threshold may assist clinical trial or experiment designers in budgetary planning, directly connecting data collection, disease progression, and costs.

We have shown how our policy pays attention to patient characteristics, such as sex, weight, severity of the disease, and knee symptoms, through more frequent follow-ups for patients at higher risk of OA. Females are generally considered to be at a higher risk of developing KOA than males~\cite{pan2016characterization,contartese2020sex}, albeit our policy selected more follow-ups for men. This observation can be partially explained by the bias in our test set or by the fact that the progression of OA may be less predictable in men and thus requires more up-to-date longitudinal information that is close to the progression event in time.

Our study still has some limitations and we acknowledge several areas for improvement. First, our method requires data from a longitudinal cohort for training. The scarcity of such data limits our ability to test with different populations. More studies are needed to incorporate the ability to learn meaningful policies from data under interventions, laying the foundations for future work. Second, our policy changes adaptively in various cost settings and might not always yield positive outcomes, especially in cases where the cost of data acquisition is high. However, in this case, other methods may not provide better results. An evaluation of acquisition cost with respect to the overall economics of disease progression ($\lambda / c$) is required. Third, our study did not address the modeling of symptomatic progression, but we recognize the potential of doing it as a part of future work. As KOA is often associated with chronic pain, integrating imaging biomarkers with symptomatic ones could lead to more precise progression outcomes and, thus, to more accurate progression modeling. However, this requires the novel definition of disease progression. Finally, our method relied on a pre-trained deep neural network (DNN) model to interpret X-ray images and extract probabilities for state input. Although we believe that this is a sufficient representation of the disease stage, other models can also be used~\cite{tiulpin2020automatic}. To shed some light on the importance of imaging data representation, and the value of this modality at all, we conducted the ablation study in \Cref{supp:imaging-imp}.

In summary, we have proposed a novel active sensing method to collect data for KOA trials and dynamically identify patients at risk of KOA progression using RL. Our method requires no human input and can run fully automatically at test time. We have demonstrated that the proposed method outperforms the existing baselines, and we have evaluated different design choices through a set of comprehensive ablation studies. The approach developed to acquire next-generation data in OA shows promise in delivering more cost-effective clinical trials and observational studies. We believe that our method may also find applications in the development of novel KOA screening and/or rehabilitation programs, or may be translated to other domains. To further advance the development of AS in KOA, we make all of our codes publicly available at \url{https://github.com/imedslab/oacostsensitivityrl}. 

\section*{Methods}
\subsection*{Reinforcement Learning with Neural Networks}
\paragraph{Markov Decision Process}
A Markov Decision Process (MDP) is defined as a tuple ($\mathbf{S}$, $\mathbf{A}$, $P$, $R$, $\gamma$), where $\mathbf{S}$ is a set of states, $\mathbf{A}$ is a set of actions. $P: \mathbf{S} \times \mathbf{S} \times \mathbf{A} \rightarrow{[0,1]}$ indicates a map defining transition probabilities between states, $R: \mathbf{S} \times \mathbf{A} \rightarrow{\mathbb{R}}$ assigns rewards to state-action pairs, and $\gamma \in [0,1]$ is the discounting factor~\cite{sutton2018reinforcement}. The decision-making policy of an agent is defined by $\pi: \mathbf{A} \times \mathbf{S} \rightarrow [0,1]$. 

In the MDP framework, the sequential decision-making process unfolds as follows~\cite{sutton2018reinforcement}. At a time step $t \in \mathbb{N}$, the environment (a ``world'' in which the agent exists and makes actions) yields a state $\mathbf{s}_t \in \mathbf{S}$ that is observed by the agent. The agent then responds with an action $a_t \in \mathbf{A}$ following the policy $\pi(a_t \mid \mathbf{s}_t)$. The policy assigns a probability to the event associated with selecting action $a_t$ at a state $\mathbf{s}_t$. Subsequently, the environment provides a reward $r_t = R(\mathbf{s}_t,a_t)$ to the agent for taking the action $a_t$. The process then repeats until the horizon is reached (i.e. $t=T$). We use the term \textit{experiences} to refer to a collection of states $\mathbf{s}_t$, next states $\mathbf{s}_{t+1}$, actions $a_t$, and rewards $r_t$ .

To measure the expected return of taking action $a_t$ at state $\mathbf{s}_t$ according to policy $\pi$, one can consider a state-action value function:
\begin{align}
\label{eq:q_function}
    Q_{\pi}(\mathbf{s}_t,a_t) = r_t + \gamma Q_{\pi}(\mathbf{s}_{t+1},a_t).
\end{align}
When solving an MDP, we aim to obtain the optimal policy $\pi^*$. In the case of Q-learning~\cite{watkins1992q},
\begin{align}
\label{eq:pi}
    \pi(\mathbf{s}) = \delta_\text{Dirac}(\argmax_{a}Q(\mathbf{s},a) = a_t^*) ~~~ \forall s \in \mathbf{S},
\end{align}
where  $\delta_\text{Dirac}(\cdot)$ is the Dirac's delta function and $a^*_t$ is the optimal action. For $\pi^*$, one can derive the Bellman equation as
\begin{align}
\label{eq:q_function_Bellman}
    Q^{*}(\mathbf{s}_t,a_t) = r_t + \gamma \max_{a_{t+1}}Q^{*}(\mathbf{s}_{t+1},a_{t+1}).
\end{align}

\paragraph{Q-function approximation}
In contemporary RL, the Q function is often approximated by a neural network (NN)~\cite{tesauro1995temporal,mnih2013playing,mnih2015human}, which we denote as $\tilde{Q}(\mathbf{s}_t,a_t, \theta)$, where $\theta$ represents the parameters of NN. The parameterized Q-function is then commonly called a Q-network. Mnih\etal\cite{mnih2013playing} introduced a method with two networks that allows us to follow the idea of Temporal Difference Learning~\cite{tesauro1995temporal}: the local Q network, approximating $Q(\mathbf{s}_t,\cdot),$ and the target Q-network, approximating $Q(s_{t+1},\cdot)$. The local Q-Network is optimized by utilizing the mean square error loss to measure the difference between the estimated Q-value and the true one, noted as:
\begin{align}
    \mathcal{L}(\theta) = \left\lVert r_t + \gamma \max_{a}Q(\mathbf{s}_{t+1},a,\theta^*) - Q(\mathbf{s}_t,a_t,\theta) \right\lVert_2^2.
\end{align}
The parameters $\theta^*$ of the target Q-Network are updated periodically using the local Q-Network to stabilize the learning process and improve convergence.
 
\paragraph{Training Q-networks}
To train a Q network from various experiences and collect information along various rollouts of an MDP, one has to balance exploration and exploitation during training. Here, we used an $\varepsilon$-greedy policy~\cite{sutton2018reinforcement} that induces the following action selection protocol. Let us draw $n_r \sim \textrm{Uniform}(0, 1)$, then 
\begin{align}
    \pi(a_t \mid \mathbf{s}_t) &= \left\{\begin{matrix*}[l]
            \arg \max_{a}Q(\mathbf{s}_t,a)  & \,\ \mathrm{if}\  n_r < \varepsilon \\
            \textrm{Random action} & \,\ \mathrm{otherwise}. \,
    \end{matrix*}
    \right.
\end{align}

Here, $\varepsilon \in [0.0,1.0]$ serves as a threshold for choosing an action, i.e., whether to follow the policy or take a risk and explore, based on a comparison with a randomly generated number in the range $[0.0,1.0]$. Typically, $\varepsilon$ is initialized at $1.0$ and gradually decays over iterations until it reaches $0.0$, indicating a transition from exploration to exploitation. The speed of this decay depends on the decay rate $d_{\varepsilon}$, such that $\varepsilon \leftarrow (1-d_{\varepsilon})\varepsilon$. 

To improve the convergence of the training procedure, the replay experience technique is usually implemented~\cite{watkins1992q,mnih2013playing,mnih2015human}. One typically stores experiences in a replay buffer so that the data that the agent collects in the past can be used for learning again and updating the network's parameters. 

\subsection*{The agent and state space} 

We designed our Q-Network to process the state revealed by the environment (i.e. multimodal input data) and predict the Q-value for each action. The input data, excluding the KOA status, were normalized across the dataset by subtracting the mean and dividing by the standard deviation. 

The state $\mathbf{s}_t$ in the quantified imaging findings, represented by a vector of probabilities extracted from a pre-trained deep neural network (DNN)~\cite{tiulpin2018automatic, nguyen2020semixup}, where each probability corresponds to a particular KOA grade~\cite{kellgren1957radiological}. As shown in~\Cref{fig:workflow}, this was done for both knees. In addition to imaging, we supplied the agent with information on patient quality of life (SF12 scores)~\cite{tarlov1989medical}, as well as symptoms via the Western Ontario and McMaster Universities Osteoarthritis Index scores (WOMAC)~\cite{bellamy1988validation}. Additional information available to the agent in the state was age, sex, and BMI, as well as the last and current data acquisition time indices. 

We first combined the KLG probability vectors from both knees and the vector of clinical variables, resulting in an input vector of $22$ elements.
The Q-Network was a simple NN, i.e., the batch norm of the inputs followed by a single hidden layer with a dropout probability of $0.2$ and sigmoid activation function, followed by a linear output layer with 2 heads. The network's output was a $2$-element vector representing Q-values of $2$ actions for each time step.

The agent was trained in an episodic setting, where each episode represented a subject's participation in an observational study. We performed the training over $1000$ epochs, each of which is a complete traverse through the whole training set. To prevent the agent from memorizing the order of the data, we shuffled them at the start of each epoch. During every episode, the agent interacted with the environment a finite number of times $T$, thereby generating experiences comprising states, actions, and rewards. These experiences were stored in a replay buffer with a capacity of $10^4$. For every $50$ episodes, we randomly sampled a batch of $256$ experiences from the replay buffer to train the network. We employed the mean square error loss and the Adam optimizer~\cite{kingma2017adam} with a learning rate of \num{5e-4}.

\subsection*{Dataset}
We conducted experiments using the Osteoarthritis Initiative (OAI) dataset, publicly available at \url{https://nda.nih.gov/oai/}. Since the OAI is a multicenter cohort, we chose one center as our test set, and the remaining data served for training purposes. To carry out our AS experiments, we included data from the baseline, 1-, 2-, 3-, and 4-year follow-ups. 

Each included participant had bilateral X-rays, and we used the method developed by~\cite{tiulpin2019kneel} to localize the knee joints~\cite{nguyen2022climat,nguyen2023clinically, tiulpin2019multimodal}. We included only those subjects whose knee images and all listed clinical data were fully available. To track disease progression, we used JSW @ 250 imaging biomarker~\cite{duryea2010comparison}, which is easily measured from radiographs and is available in OAI. The KLG scores, representing the severity of KOA disease, were computed from a DL-based model~\cite{nguyen2020semixup}.  Furthermore, we utilized clinical data including age, sex, BMI, physical SF12 score~\cite{ware199612}, past injury records, and past surgery records. To aid the models with the status of the joint function, stiffness, and pain, we also employed the total WOMAC score~\cite{bellamy1988validation} in our feature vector. After data curation, we obtained radiographs and clinical variables from a total of $1620$ participants in the OAI. The data statistics at the baseline are summarized in~\Cref{tbl:data_chracteristic}.

\begin{table}[ht!]
\caption{Characteristics of the OAI participants at the baseline. $(\star)$ Participants were classified as ``Asymptomatic'' if they obtained a WOMAC score of $0$, and ``Symptomatic'' otherwise. $(\star\star)$ Participants' KOA severity was classified based on the Kellgren-Lawrence grading (KLG) system at the baseline. Those with both knees obtained KLG 0/1 are labeled as ``Healthy'', while individuals with at least one knee graded KLG 2, KLG 3, or KLG 4 are categorized as ``Mild'', ``Moderate'', or ``Severe'' KOA, respectively.}
\centering
\scalebox{0.9}{
\begin{tabular}{llrr}
\toprule
\textbf{Category} &  \textbf{Sub-groups} & \textbf{Training} & \textbf{Test} \\
\midrule \midrule
Subjects & - & 1199 & 421 \\
Knees & - & 2398 & 842 \\
\midrule
\multirow{2}{*}{Age} & < 55  & 345 (28.8\%)    &    103 (27.6\%)   \\
 & 54-64      &   404 (33.7\%)  &   134 (31.8\%)         \\
 & 65-74      &   359 (29.9\%)  &   146 (34.7\%)        \\
  & > 75      &   91 (7.6\%)  &   38 (9.0\%)         \\
\midrule
\multirow{2}{*}{Sex} & Male  &    488 (40.7\%)    &  180 (42.8\%)          \\
& Female & 711(59.3\%) & 241 (57.2\%) \\
\midrule
\multirow{4}{*}{BMI} & Underweight & 2 (0.17\%) & 1 (0.24\%) \\
&Healthy & 248 (20.7\%) & 98 (23.28\%) \\
& Overweight & 471 (39.3\%) & 185 (43.9\%) \\
& Obesity & 478 (39.7\%) & 137 (32.5\%) \\
\midrule
\multirow{2}{*}{Symptomatic$^{(\star)}$}    &   Yes          &  1006 (83.9\%)        &   340 (80.8\%)                      \\
&No & 193 (16.1\%) & 81(19.2\%) \\
\midrule
\multirow{4}{*}{Severity$^{(\star\star)}$}    &   Healthy          &  344 (38.7\%)        &   120 (28.5\%)                      \\
&Mild & 452 (37.7\%) & 171 (40.6\%) \\
&Moderate & 345 (28.8\%) & 109 (25.9\%) \\
&Severe & 58 (4.8\%) & 21 (5.0\%) \\
\bottomrule
\end{tabular}
}
\label{tbl:data_chracteristic}
\end{table}

\subsection*{Defintion of Multi-progression}

In this section, we introduce a procedure to construct a ground truth set of progression events for individual subjects using the decrease in the fixed medial JSW (fJSW), a robust and effective biomarker to capture structural degeneration in KOA~\cite{ratzlaff2018quantitative,duryea2010comparison}. We use $\phi_L(\zeta)$ and $\phi_R(\zeta)$ to respectively denote the measured fJSWs of the left and the right knees at every observational time point $\zeta \in \{0, \dots, T\}$. Changes in the left and right fJSWs at time point $\zeta$ relative to a reference time point $\zeta_r$, where $\zeta_r < \zeta$, are denoted by $d_L(\zeta_r, \zeta)$ and $d_R(\zeta_r, \zeta)$, respectively.
The following procedure aims to generate the set of progression events $\mathbf{M}_p$, which was done once before training the RL policy.

Initially, we set the first reference time point to the baseline examination, denoted by $\zeta_r=0$, and let $\mathbf{M}_p$ be empty. At each time step $\zeta\in \{1,\dots, T\}$, we calculate $d_L(\zeta_r, \zeta)$ and $d_R(\zeta_r, \zeta)$ 
as follows:
\begin{align}
    \label{eq:reward_dL} d_L(\zeta_r, \zeta) &= \max(0, \phi_L(\zeta_r) - \phi_L(\zeta)), \\ 
    \label{eq:reward_dR} d_R(\zeta_r, \zeta) &= \max(0, \phi_R(\zeta_r) - \phi_R(\zeta)). 
\end{align}
Next, we define the patient-level KOA worsening at $\zeta$ relative to $\zeta_r$ as:
\begin{equation}
    d_{LR}(\zeta_r, \zeta) = \max(d_L(\zeta_r, \zeta), d_R(\zeta_r, \zeta)).
\label{eq:patientL_JSW_changes}
\end{equation}

If $d_{LR}(\zeta_r, \zeta)$ exceeds a threshold $\kappa$, a progression event is recorded, and we add $\zeta$ to $\mathbf{M}_p$. We then set $\zeta$ as the new reference time point (i.e., $\zeta_r:=\zeta$) and continue the procedure at $\zeta+1$. If the threshold $\kappa$ is not exceeded, the procedure also proceeds to $\zeta+1$. Ultimately, when $\zeta$ reaches $T$, we terminate the process and obtain the progression set $\mathbf{M}_p$ containing the time points of all the progression events. In our experiments, we set $\kappa$ to 0.7mm~\cite{hunter2023biomarkers} to account for the imperfect reliability of the biomarker. The details on the progression happening at a patient level resulted in data shown in \Cref{tbl:JSW_progression}.

\subsection*{The cost of multi-joint disease progression} The biggest challenge we had to solve when designing the reward is the harmonization of two semantically distinct units, bilateral structural changes (measured in $\mathrm{mm}$) and acquisition costs (measured in \$) -- in our reward function. Using domain knowledge, we propose a novel approach to convert the reduction of fJSW to monetary value. Specifically, we assume that (1) the end stage of the disease is TKR and (2) if KOA is detected early, it can presumably be slowed~\cite{yao2023osteoarthritis}. From these assumptions, we define $c$ as the monetary loss of $1 \mathrm{mm}$ of articular cartilage (MLAC), calculated as:
\begin{align}    
    c &= \frac{\mathrm{TKR \ cost}\ (\$) }{\mathrm{Mean\ \ JSW\ of\ healthy\ knees}\ (\mathrm{mm})},
\label{eq:cost_convert}
\end{align}
where the mean fJSW can be computed from an observational cohort (e.g., OAI), and the TKR cost is pre-defined for a healthcare system where the RL method is deployed.
To this end, we can map bilateral structural changes (see eq. \eqref{eq:patientL_JSW_changes}) into financial costs as
\begin{equation}
    \delta(\cdot,\cdot) = cd_{LR}(\cdot,\cdot).
\end{equation}
\subsection*{Hyperparameters of the reward function}
We utilized the United States Current Procedural Terminology (CPT) system to search for the expenses associated with medical imaging and TKR~\cite{thorwarth2004concept,hirsch2015current}. Based on these data, TKR surgery (CPT code: 27446) can reach \$50,000 per patient. A follow-up visit consisting of consultant fees (CPT code: 73560) and knee X-ray imaging (CPT codes: 73560, 73562, 73564, 73565) may range from \$300 to \$1,000. The medical cost varies depending on region, insurance, and other caring services. In our experiments, we empirically set the cost of TKR at \$10,000, and the cost of each hospital visit at \$500. 

The average healthy JSW was set to $5\,\mathrm{mm}$ according to~\cite{buckland1995joint,anas2013digital}. We compute $c=\$10,000/5\,\mathrm{mm}=2000 \, \$/\mathrm{mm}$. For numerical stability, we scaled all monetary hyperparameters by $1000$. 

We empirically chose $\alpha=0.5$ as a parameter to reduce the negative reward associated with late visits and set $\beta=0.3c$ to assign positive rewards for correct dismissal. $\lambda=0.5$ was used in all our experiments and we also conducted an ablation study to validate how our method performs in different settings. Hyperparameter $\kappa$ was set to

\subsection*{Evaluation} 
We validated the reference  policies using the Reward Per Person over $4$ years (RPP), formulated as
\begin{equation}
  \bar{R} = \frac{1}{N}\sum_{i=1}^{N} \sum_{t=0}^{T} r^{(i)}_t,
  \label{eq:ARPP}
\end{equation}
where $N$ is the number of subjects. RPP represents the monetary utility that each subject with KOA can earn or overspend on AS. To quantitatively evaluate the correctness of the policies, we calculated the balanced accuracy and recall scores at each time step and averaged them across all the follow-ups. We repeated our experiments using $10$ different random seeds and reported the average results along with their standard errors across the runs.

\section*{Acknowledgements}

Research was supported by the Research Council of Finland (6GESS profiling program, project 336449), the Sigrid Juselius Foundation,  University of Oulu strategic funding and the European Union Horizon Program (STAGE project, decision 101137146). The project was also supported by the Finnish Doctoral Program Network in Artificial Intelligence, AI-DOC (decision number VN/3137/2024-OKM-6). CSC – IT Center for Science, Finland is kindly acknowledged for providing generous computational resources. 

\section*{Author contributions statement}

K.N. conducted the research and wrote the initial draft of the manuscript. PP contributed to the methodology,  evaluation approach, and draft of the manuscript. H.H.N. contributed to the methodology and the draft of the manuscript. A.T. conceived the idea for the study, acquired funding, led the research, and finalized the draft of the manuscript. All authors contributed to the review and final approval of the manuscript.

\section*{Additional information}
The authors have no competing interests. The funding sources did not play any role in the design or execution of this study.

\bibliography{refs}
\supplementarysection
\subsection{Reference methods details}\label{supp:baselines}
The policies used as the reference methods are described below.
\begin{itemize}
\itemsep0pt
    \item \textbf{Random}. Randomly chooses an action with different probabilities at every time point for every patient.
    \item \textbf{NOS}. Dismissal/follow-up skip actions are taken to all subjects at all follow-up points.
    \item \textbf{Biannual sensing (BANS)}. Recommends following up on every subject every 2 years.
    \item \textbf{Annual sensing (ANS)}. Recommends following up on every subject annually.

    \item \textbf{LR} ~\cite{berkson1944application}. We generated all possible input states with the different time index combinations for the training set. The target is in a set of $\{0,1\}$. We fit these unroll data to the LR model. By grid scanning the model parameters, the best LR policy was achieved. Next, we tested the trained LR policy at each time point respectively. The input for the next time point was modified according to predictions, and the update rule followed the definition of multi-progression.
    \item \textbf{Cox regression} (Cox)~\cite{cox1972regression}. Time-to-event method that allows to estimate when the progression event occurs. This method aims to detect the first KOA progression based on only the baseline data. Hence, after achieving the best model through grid searching parameters, to earn the multi-events predictions, we adjusted the way of testing the model. In testing, based on the predicted progression event, new inputs were updated, and we requested the model to continuously forecast the next event until the prediction exceeds the evaluated time $T$.
    \item \textbf{GRU}~\cite{chung2014empirical} \textbf{and LSTM}~\cite{hochreiter1997long}. RNN-based methods to estimate a sequence from input. We implemented dynamic learning into both GRU and LSTM.
    Dynamic learning allows us to make predictions based on the data in the latest visit. We applied an action-dependent mask for every input state during the training process of GRU and LSTM. The combination of the new input state and the hidden state from the previous step was implemented into models to estimate the next action.

    \item \textbf{Dynamic-DeepHit} (DeepHit)~\cite{lee2019dynamic} incorporates the longitudinal data and learns the time-to-event distributions by utilizing DL, aiming to update survival predictions dynamically. DeepHit proposed 2 subnetworks: a shared subnetwork to learn the history of measurements; and a cause-specific subnetwork to capture the risk for each competing event. The former network consists of an RNN structure and an attention mechanism, while the latter one is composed of fully-connected layers. This method dynamically updated the risk of progression when new data recorded. We implemented their published repository with paper and prepared unrolled data for the training process. 
\end{itemize}

\subsection{Detailed data}\label{supp:result}

We show all the data behind the figures in the manuscript in \Cref{tab:bacc-climat,tab:fig4,tab:number-of-visits}. Statsitics on the number of progressors at every follow-up are shown in~\Cref{tbl:JSW_progression}.

\begin{table}[ht!]
\centering
\caption{Reward per person (RPP) achieved by baselines and our policy in various cost settings. In subfigure (a), adjusting follow-up cost $\lambda$ while keeping constant monetary cost of a 1mm loss in articular cartilage (MLAC) $c=2$. In subfigure (b), varying $c$ while $\lambda=0.5$. In subfigure (c), performance of our policy in full factorial of $\lambda$ and $c$ parameters.}\label{tab:fig4}
\subfloat[With $c=2$, varying $\lambda$]{
\begin{tabular}{ccccc}
\toprule
 \textbf{$\lambda$} & \textbf{NOS} & \textbf{ANS} & \textbf{CLIMATv2} & \textbf{Ours} \\
\midrule
    0.00 & -1.64 & 1.14 & -0.51 & 1.18 \\ 
    0.25 & -1.64 & -0.06 & -0.74 & 0.54 \\
    0.50 & -1.64 & -0.86 & -0.89 & 0.34 \\
    0.75 & -1.64 & -2.06 & -1.12 & -0.23 \\
    1.00 & -1.64 & -2.86 & -1.27 & -0.69 \\
    1.50 & -1.64 & -4.86 & -1.65 & -1.10 \\
    2.00 & -1.64 & -6.86 & -2.03 & -1.16 \\
\bottomrule
\end{tabular}
\label{tab:fig4a}
}
\subfloat[With $\lambda=0.5$, varying $c$]{
\begin{tabular}{ccccc}
\toprule
 \textbf{$c$} &\textbf{NOS} & \textbf{ANS} & \textbf{CLIMATv2} & \textbf{Ours} \\
\midrule
    0.50 & -0.41 & -1.71 & -0.51 & -0.36 \\
    1.00 & -0.82 & -1.43 & -0.63 & -0.35 \\
    1.50 & -1.23 & -1.15 & -0.76 & -0.05 \\
    2.00 & -1.64 & -0.87 & -0.89 & 0.34 \\
\bottomrule
\end{tabular}
\label{tab:fig4b}
}

\subfloat[Reward-per-person (RPP) gained by our reinforcement learning-based policy depending on $c$ and $\lambda$ combinations. Underscored numbers indicate positive or close to $0$ RPP values.]{
\begin{tabular}{ccccccc}
\toprule
\multirow{2}{*}{\textbf{$\lambda$}}&\multicolumn{6}{c}{\textbf{$c$}} \\
 \cmidrule{2-7}
 & \textbf{0.50} & \textbf{1.00} & \textbf{1.50} & \textbf{2.00} & \textbf{3.00}  & \textbf{4.00}  \\
\midrule
 \textbf{0.00} & \underline{0.31}  & \underline{0.61}  & \underline{0.92}  & \underline{1.18}  & \underline{1.85}  & \underline{2.45}  \\
\textbf{0.25} &-0.14 & \underline{0.15}  & \underline{0.36}  & \underline{0.54}  & \underline{1.19}  & \underline{1.83}  \\
\textbf{0.50} &-0.36 & -0.36 & \underline{-0.05} & \underline{0.34}  & \underline{0.65}  & \underline{1.20}  \\
 \textbf{0.75} &-0.44 & -0.54 & -0.48 & -0.23 & \underline{0.16}  & \underline{0.71}  \\
 \textbf{1.00} &-0.49 & -0.65 & -0.75 & -0.69 & \underline{-0.09} & \underline{0.16}  \\
 \textbf{1.50} &-0.45 & -0.83 & -0.97 & -1.10 & -1.01 & -0.52 \\
 \textbf{2.00} &-0.42 & -0.90 & -1.16 & -1.16 & -1.49 & -1.35 \\
 \textbf{2.50} &-0.41 & -0.85 & -1.31 & -1.45 & -1.61 & -1.85 \\
\textbf{3.00} &-0.41 & -0.85 & -1.22 & -1.75 & -1.68 & -2.09 \\
 \textbf{3.50} &-0.41 & -0.83 & -1.21 & -1.78 & -2.01 & -2.18 \\
 \textbf{4.00} &-0.41 & -0.82 & -1.28 & -1.78 & -2.22 & -2.33 \\

\bottomrule
\end{tabular}
\label{tab:fig4c}
}

\end{table}

\begin{table}[t]
\caption{Number of visits (mean and standard error) of $3$ policies according to pairs of Kellgren-Lawrence grades (KLGs) at the baseline. Those policies were trained with different follow-up costs hyper-parameter $\lambda=\{0.5,0.8,1.0\}$.}\label{tab:number-of-visits}
\centering
\begin{tabular}{cccc}
\toprule
\multirow{2}{*}{\textbf{Baseline KLGs}}& \multicolumn{3}{c}{\textbf{Policy}} \\
\cmidrule{2-4}
     & \textbf{$\lambda=0.5$} & \textbf{$\lambda=0.8$} & \textbf{$\lambda=1.0$} \\
\midrule
(0, 0)  & 1.33 $\pm$ 0.18 & 1.07 $\pm$ 0.10 & 1.03 $\pm$ 0.06 \\
(0, 1)  & 1.34 $\pm$ 0.20 & 1.13 $\pm$ 0.11 & 1.04 $\pm$ 0.07 \\
(1, 1)  & 1.65 $\pm$ 0.29 & 1.31 $\pm$ 0.22 & 1.14 $\pm$ 0.13 \\
(0, 2)  & 1.94 $\pm$ 0.32 & 1.46 $\pm$ 0.26 & 1.18 $\pm$ 0.15 \\
(1, 2)  & 2.01 $\pm$ 0.33 & 1.52 $\pm$ 0.23 & 1.31 $\pm$ 0.18 \\
(2, 2)  & 2.06 $\pm$ 0.33 & 1.59 $\pm$ 0.24 & 1.25 $\pm$ 0.16 \\
(0, 3)  & 3.03 $\pm$ 0.31 & 2.31 $\pm$ 0.30 & 1.51 $\pm$ 0.20 \\
(1, 3)  & 2.83 $\pm$ 0.32 & 2.20 $\pm$ 0.30 & 1.53 $\pm$ 0.20 \\
(2, 3)  & 2.96 $\pm$ 0.32 & 2.16 $\pm$ 0.29 & 1.76 $\pm$ 0.26 \\
(3, 3)  & 3.61 $\pm$ 0.21 & 2.88 $\pm$ 0.28 & 2.15 $\pm$ 0.25 \\
(1, 4)  & 3.83 $\pm$ 0.12 & 3.30 $\pm$ 0.23 & 2.02 $\pm$ 0.21 \\
(2, 4)  & 3.38 $\pm$ 0.17 & 2.88 $\pm$ 0.25 & 2.28 $\pm$ 0.23 \\
(3, 4)  & 3.96 $\pm$ 0.06 & 3.24 $\pm$ 0.28 & 2.24 $\pm$ 0.27\\
\bottomrule
\end{tabular}
\end{table}

\begin{table}[ht!]
\centering
\caption{Balanced accuracy (BA) and recall scores (mean and standard error, \%) of our method and CLIMATv2 performed with knee-level and patient-level approaches, presented at each year. Bold numbers represent the highest scores among all methods.}\label{tab:bacc-climat}
\begin{tabular}{llcll}
\toprule
  & \textbf{Method} & \textbf{Year(s)} & \textbf{BA(\%)} & \textbf{Recall(\%)} \\
\midrule
 \parbox[t]{2em}{\multirow{8}{*}{\rotatebox[origin=c]{90}{Knee-level}}}  & \multirow{4}{*}{CLIMATv2} & 1 & 57.09 $\pm$ 1.69 & 24.29 $\pm$ 6.22\\
&& 2 & 50.87 $\pm$ 1.26 & 23.77 $\pm$ 4.30 \\
&& 3 & 56.34 $\pm$ 0.84 & 46.15 $\pm$ 5.00 \\
&& 4 & 55.60 $\pm$ 1.16 & 38.59 $\pm$ 4.55 \\
\cmidrule{2-5}
& \multirow{4}{*}{Ours} & 1 & 60.43 $\pm$ 0.72 & 58.98 $\pm$ 3.84 \\
&& 2 & \textbf{59.26 $\pm$ 0.57} & 82.92 $\pm$ 1.83 \\
&& 3 & 47.30 $\pm$ 0.80 & 32.88 $\pm$ 0.72 \\
&& 4 & 59.41 $\pm$ 0.73 & 38.51 $\pm$ 1.95 \\
\midrule
\parbox[t]{2em}{\multirow{8}{*}{\rotatebox[origin=c]{90}{Patient-level}}} & \multirow{4}{*}{CLIMATv2} & 1 & 51.83 $\pm$ 0.77 & 7.55 $\pm$ 3.18 \\
&& 2 & 56.71 $\pm$ 0.95  & 42.83 $\pm$ 5.97 \\
&& 3 & 58.73 $\pm$ 0.79  & 50.58 $\pm$ 3.24 \\
&& 4 & 58.38 $\pm$ 0.89 & 50.16 $\pm$ 4.80 \\
\cmidrule{2-5}
& \multirow{4}{*}{Ours} & 1 & \textbf{62.81 $\pm$ 0.57} &  \textbf{74.08 $\pm$ 2.62} \\
&& 2 & 53.56 $\pm$ 0.44 & \textbf{96.98 $\pm$ 0.85} \\
&& 3 & \textbf{60.40 $\pm$ 0.67} & \textbf{67.69 $\pm$ 1.03} \\
&& 4 &  \textbf{64.55 $\pm$ 0.74} & \textbf{55.94 $\pm$ 1.56} \\
\bottomrule
\end{tabular}

\label{tab:fig5}
\end{table}

\subsection{Importance of imaging data}\label{supp:imaging-imp}
We also evaluated the impact of imaging data on the performance of the proposed method. To set up a benchmark, we conducted experiments without the contributions of radiographs, thereby excluding the KLG probabilities from the agent's input. We then incorporated one of the three distinct types of KLGs into our experiments. The first type, referred to as the ``KLG ground truth'' corresponds to the manual grades produced by the radiologists. The second and third types named the ``image features'' and the ``KLG probabilities'', respectively, were extracted from a DNN that had been trained to predict the KLGs from radiographs automatically~\cite{tiulpin2018automatic,nguyen2020semixup}.

\Cref{tbl:result_input_type} illustrates the performance of our RL-based method when the image-derived data type changes. Specifically, the setup in which the imaging data were eliminated from the state completely, resulted in poor performance with a negative reward of $-0.05$ and an average BA of $55.02\%$. Meanwhile, using ground-truth KLG in states led to substantial improvements of $0.20$ average reward and $2.67\%$ average BA. When we used image features extracted from the network instead of the KLGs, we obtained a negative reward of $-0.86$ and performance drops with an average BA of $51.91\%$ and an average recall score of $26.62\%$. In contrast, having KLG probabilities in the state yielded a positive total reward of $0.20$, and outperformed other setups with the average recall of $73.67\%$.

\begin{table}[ht!]
\caption{Performance of our reinforcement learning-based method on the deployment set with various types of radiographic findings. The reference model is a setting without images. The remaining experiments involved manual Kellgren-Lawrence grades (KLGs) and automatic KLGs, serving as knee osteoarthritis findings. Automatic KLGs were extracted from a pre-trained deep neural network under ``Image features'' or ``KLG probabilities''. BA = balanced accuracy.
}
\label{tbl:result_input_type}
\centering
\scalebox{0.85}{
\begin{tabular}{llllll}
\toprule
\textbf{Input}   & \textbf{Reward} & \textbf{BA} & \textbf{Recall} \\
\midrule
Without imaging & -0.05 $\pm$ 0.03 & 55.02 $\pm$ 0.30 & 66.87 $\pm$ 1.53 \\
\midrule
KLG ground truth      &   0.15 $\pm$ 0.02           &  57.69 $\pm$ 0.27        &   64.53 $\pm$ 1.33                       \\
\midrule
Image features & -0.86 $\pm$ 0.03               & 51.91 $\pm$ 0.41       & 26.62 $\pm$ 1.16           \\
KLG probabilities & 0.20 $\pm$ 0.03      & 60.33 $\pm$ 0.32       & 73.67 $\pm$ 1.00             \\

\bottomrule
\end{tabular}
}
\end{table}

\begin{table}[ht!]
\caption{The statistics of subjects progressed with knee osteoarthritis in the training and test sets after the baseline examination.
}
\centering
\scalebox{0.83}{
\begin{tabular}{lcccc}
\toprule
\textbf{Year(s)} & \textbf{1} & \textbf{2} & \textbf{3} & \textbf{4} \\
\midrule
Training set & 161 (13.4\%) & 211 (17.6\%) & 242 (20.2\%) & 233 (19.4\%) \\
Test set & 49 (11.6\%) & 53 (13.6\%) & 52 (12.4\%) & 64 (15.2\%) \\
\bottomrule
\end{tabular}
}
\label{tbl:JSW_progression}
\end{table}

\end{document}